# Probing the limit of hydrologic predictability with the Transformer network

Jiangtao Liu[1], Yuchen Bian[2] and Chaopeng Shen*,[1]
[1] Civil and Environmental Engineering, The Pennsylvania State University
[1] Amazon Search Science and AI
* Corresponding author: Chaopeng Shen, cshen@engr.psu.edu

Abstract

For a number of years since its introduction to hydrology, recurrent neural networks like long short-term memory (LSTM) have proven remarkably difficult to surpass in terms of daily hydrograph metrics on known, comparable benchmarks. Outside of hydrology, Transformers have now become the model of choice for sequential prediction tasks, making it a curious architecture to investigate. Here, we first show that a vanilla Transformer architecture is not competitive against LSTM on the widely benchmarked CAMELS dataset, and lagged especially for the high-flow metrics due to short-term processes. However, a recurrence-free variant of Transformer can obtain mixed comparisons with LSTM, producing the same Kling-Gupta efficiency coefficient (KGE), along with other metrics. The lack of advantages for the Transformer is linked to the Markovian nature of the hydrologic prediction problem. Similar to LSTM, the Transformer can also merge multiple forcing dataset to improve model performance. While the Transformer results are not higher than current state-of-the-art, we still learned some valuable lessons: (1) the vanilla Transformer architecture is not suitable for hydrologic modeling; (2) the proposed recurrence-free modification can improve Transformer performance so future work can continue to test more of such modifications; and (3) the prediction limits on the dataset should be close to the current state-of-the-art model. As a non-recurrent model, the Transformer may bear scale advantages for learning from bigger datasets and storing knowledge. This work serves as a reference point for future modifications of the model.

## *Introduction*

Rainfall-runoff modeling is essential for flood prediction, water resource management, and environmental protection (Hrachowitz & Clark, 2017). Rainfall-runoff modeling is a critical aspect of hydrology, as it models the intricate relationships between precipitation, watershed characteristics, and streamflow. The introduction of long short-term memory (LSTM) networks marked a significant advancement in this field for numerous variables of interest including soil moisture (Fang & Shen, 2017; Feng et al., 2020), streamflow (Feng et al., 2020, 2021; Kratzert et al., 2019; Xiang & Demir, 2020), water temperature (Rahmani, Lawson, et al., 2021; Rahmani, Shen, et al., 2021), and groundwater levels (Afzaal et al., 2020; Wunsch et al., 2022). For these applications, LSTM consistently outperformed traditional models such as the autoregressive integrated moving average (ARIMA) and support vector machines (SVM) (Papacharalampous et al., 2018). The LSTM's ability to learn many-step dependencies and handle variable-length input sequences has proven particularly advantageous in capturing the inherent complexity and non-stationarity of hydrological processes (Hochreiter & Schmidhuber, 1997).

As a recurrent neural network (RNN), LSTM has to step through time steps one by one, accumulate changes to states after each step, and apply the neural networks many times, which lead to some limitations. The recurrent nature lends RNNs to an issue called vanishing gradient (Hochreiter, 1991; Hochreiter et al., 2001), where the gradient of the loss with respect to the network weights is very small, making their training extremely slow. This issue limits the length of the training sequence, and reduces the impact of inputs in the longer-term past on present predictions. This could be one of the reasons why baseflow was previously identified as a limitation (Feng et al., 2020). Even though LSTM was developed to mitigate this issue and can suppress it better than the original RNNs, it is not immune to it. Furthermore, recurrence means these time steps must be taken in sequence --- the time steps cannot be run in parallel. This poses a limitation in parallel efficiency and thus the scale of data that can be trained on.

In most applications outside hydrology, the transformer architecture (Vaswani et al., 2017) has demonstrated superior performance over LSTM networks in various domains, including machine translation, speech recognition (Karita et al., 2019), natural language processing and sentiment analysis (Devlin et al., 2019), question answering (Rajpurkar et al., 2018), computer vision (Carion et al., 2020), protein structure prediction (Rives et al., 2021), and music generation (Huang et al., 2018). Transformer model uses the attention mechanism, where each word (or "input token") is transformed into three different kinds of information: a 'query' that asks how relevant other words are to it, a 'key' that responds to others' queries about its relevance, and a 'value' that carries the word's actual meaning. The model calculates the relevancies between the query and keys of all words, then combines the values of the most relevant words to understand the current word better. With LSTM, the most recent input tokens are always more important than further-away ones, whereas a Transformer could learn to put more focus to further-away tokens (Dehghani et al., 2019; Raganato & Tiedemann, 2018), which makes it ideal for language modeling. Moreover, as it does not have recurrence, a Transformer can run the time steps in parallel and can scale up in parallel computation when more data and more GPUs are available. Considering such success, there should be a heightened interest in harnessing transformers for hydrologic applications. However, only a few studies have employed transformers in hydrology (Li & Yang, 2019; Xu et al., 2021) (which focused on near-term forecast), and no work reported its results on standardized, well-understood benchmark problems like the Catchment Attributes and Meteorology for Large-sample Studies (CAMELS) dataset (Addor et al., 2017, 2017; Newman et al., 2014). It is then intriguing if its advantages over recurrent networks apply to the natural systems, which lacked the irregular sequential structure commonly found in languages.

While some past studies have claimed some architectures' superior performance compared to LSTM, most of the time the conclusions were highly conditional on using a small dataset for benchmarking (Abed et al., 2022; Amanambu et al., 2022; Ghobadi & Kang, 2022), or using procedures and configurations, e.g., training and test periods, sites, and forcing data, different from published benchmarks (Yin et al., 2022, 2023), or on a case study which were not tested independently by other teams (Koya & Roy, 2023; Liu et al., 2022). In the interest of reproducibility and comparability which underpin scientific progress, it is a good idea to benchmark under the same conditions, on the same (reasonably large) dataset. In addition, data-driven deep learning models enjoy the feature of "data synergy", where larger and more diverse data leads to stronger and more robust models (Fang et al., 2022; Kratzert et al., 2020; Pasquiou et al., 2022; Yang et al., 2023). Thus small-data comparison results may be no longer valid for the case with more data. Thus far, on the CAMELS dataset (Addor et al., 2017; Newman et al., 2014), both Kratzert et al. (2019) and Feng et al., (2021) reported very comparable metrics with LSTM --- 0.72 for 571 basins with the NLDAS forcing alone, making this a reliable benchmark that is so far not exceeded by other models. Furthermore, Kratzert simultaneously employed multiple forcing dataset (NLDAS, Maurer and Daymet) for LSTM and obtained a Kling-Gupta model efficiency coefficient (KGE) of 0.80, which is the record on this dataset.

In this study, we investigate the performance of the Transformer architecture in rainfall-runoff modeling compared to LSTM, using the CAMELS dataset. We analyze the performance of single models and ensembles for both architectures and examine the models' ability to handle multiple forcings and mixed forcing cases. Even though we had expected the difference to be small compared to LSTM, we aim at establishing a reference point where future studies can contrast and compare to. Our findings contribute to the understanding of the strengths and limitations of both LSTM and Transformer models in hydrological predictions and highlight the potential of the Transformer as an alternative and scalable solution for hydrologic modeling.

## *Results and Discussion*

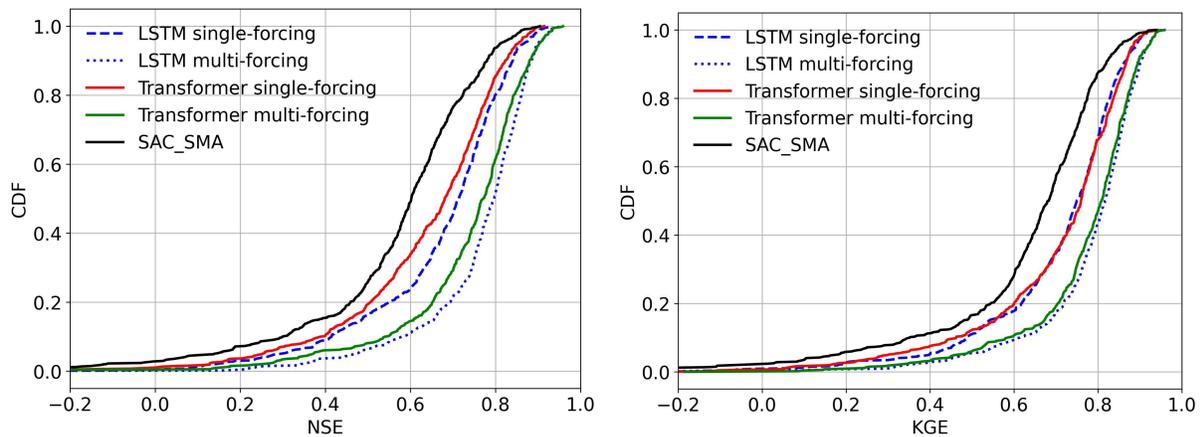

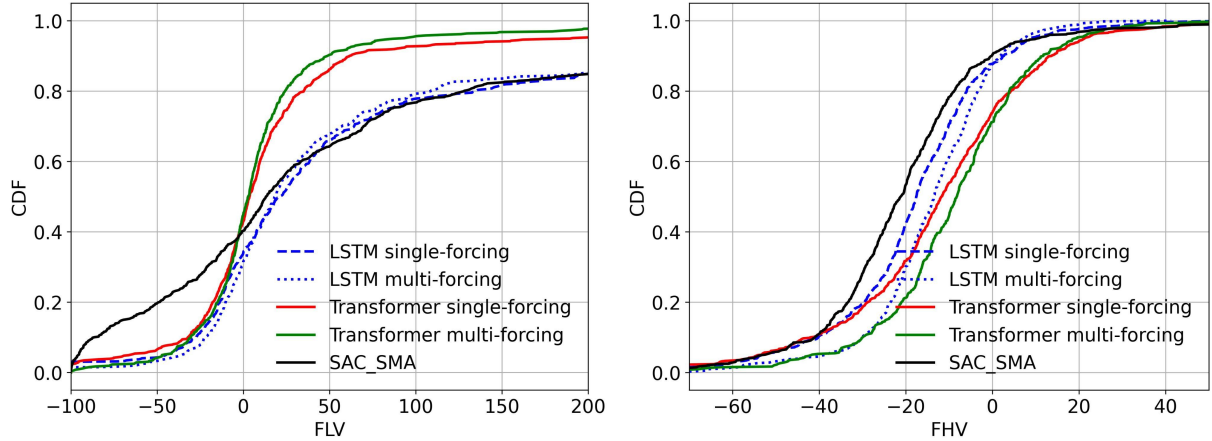

*Figure 1. Comparative analysis of Cumulative Density Function(CDF) across various models including LSTM, Transformer, and SAC-SMA, with unit in mm/day. The model encompass single and multi-forcing data for models. The training period ranged from 1 October 1999 to 30 September 2008, while the testing period ranged from 1 October 1989 to 30 September 1999. The figure depict the following comparisons: (a) NSE vs CDF, (b) KGE vs CDF, (c) FLV vs CDF, and (d) FHV vs CDF. Single-forcing models were implemented on a set of 671 basins, whereas multi-forcing models were applied to a subset of 531 basins.*

*Table 1. Comparative performance metrics for single and multi-forcing experiments of LSTM, vanilla transformer, and modified transformer models. Training dates for the models span from 1 October 1999 to 30 September 2008, while testing dates cover 1 October 1989 to 30 September 1999. We conducted an evaluation of single forcing on 671 basins and multi-forcing on 531 basins, employing the LSTM model results from Kratzert et al., 2019, originally evaluated on 531 basins. To broaden our insights into the impacts of single-forcing on the entire dataset and a fair comparison, we retrained their model on an expanded set of 671 basins with single NLDAS dataset. These numbers are only very slightly different from Kratzert et al., 2019. The means for NSE, KGE, FHV, and FLV are averages from the 10 different ensemble members, each with a different random seed, while the standard deviations for FHV and FLV are calculated for the ensemble members.*

|  | NLDAS | | | Multi-forcing | | |
|---|---|---|---|---|---|---|
|  | LSTM | Vanilla Transformer | Modified Transformer | LSTM | Vanilla Transformer | Modified Transformer |
| KGE (mean) | 0.73 ±0.003 | 0.71 ±0.007 | 0.74 ±0.007 | 0.80 ±0.004 | 0.77 ±0.016 | 0.80 ±0.007 |
| FHV (mean) | -17.49 ±0.58 | -26.66 ±2.83 | -18.00 ±2.94 | -11.91 ±0.549 | -21.54 ±2.64 | -9.19 ±4.01 |
| FLV (mean) | -2.82 | 3.31 | 2.28 | 2.57 | 0.77 | 2.72 |

| | ±8.15 | ±2.34 | ±4.24 | ±4.072 | ±1.65 | ±2.41 |

For the single-forcing CAMELS benchmark (671 basins), the vanilla Transformer is outperformed by LSTM (Table 1; Figure 1). Overall, the vanilla Transformer fell behind LSTM in all metrics, although not by much. Since KGE is a reliable performance metric, we choose to focus on this metric (Knoben et al., 2018, 2019), for which the vanilla Transformer reported 0.71, compared to 0.73 for the LSTM. These results suggest that, without modification, the vanilla Transformer is missing some critical ability to simulate hydrologic processes.

The vanilla Transformer's under-performance is a curious case as it has been widely recognized that "attention is all you need" (Vaswani et al., 2017) in sequential modeling, and we have several interpretations. First, it is possible that the data size is too small here and Transformer's advantage would emerge for larger quantities of data. Second, the natural hydrologic process is a "Markovian" system where the states at a time, rather than more remotely-in-the-past steps, completely determine the system's trajectory for the future time steps along with the forcings. To be more concrete, the soil moisture today, rather than any previous days', would have far more impact on tomorrow's discharge. This is in strong contrast to human languages where the order of the words can often be inverted, which would favor the attention-based Transformer architecture. Third, the accumulation of moisture and its nonlinear interactions makes memory effects important, while the Transformer does not have memory and is not necessarily strong at capturing the effects of memory. Regardless of the reason, the results mean that the vanilla Transformer is not optimal and further changes are needed to model the natural systems.

The modified Transformer, on the other hand, shows very similar median metrics as LSTM while the variabilities among ensemble members are different. Its KGE (0.74) is slightly higher than LSTM (0.73), and the differences in FLV and FHV from LSTM's values are too small to call an advantage considering their variabilities. The ensemble standard deviation of KGE is 0.003 with LSTM and 0.007 with the modified Transformer. We notice that the LSTM has smaller ensemble standard deviation FHV than the modified Transformer, while the opposite is true for FLV. The ensemble standard deviation of median FHV is 0.58 for LSTM and 2.94 for the modified Transformer. That value for the FLV is 8.15 for the LSTM and 4.24 for the modified Transformer. This suggests while we obtain very similar overall metrics, the LSTM and the modified Transformer preferentially address different parts of the hydrograph. LSTM more reliably focus on the high-flow regime (quantified by the ensemble standard deviation of FHV) than the modified Transformer but the latter can more reliably capture the long-term dependence (quantified by ensemble standard deviation of FLV). It seems there is some tradeoff for the different flow regimes.

The multiforcing experiment generally shows similar patterns: the vanilla Transformer falls behind the other two models, which have very similar ensemble-mean performance metrics but different ensemble standard deviations. The high KGE (0.80) and slightly better-than-LSTM FHV (9.19) for the modified Transformer demonstrates that it, too, is able to fuse different forcing dataset as just LSTM. Just as the NLDAS experiment, the modified Transformer has a larger stochastic variability (quantified by ensemble standard deviation) FHV but smaller variability for FLV. Because both FHV and FLV have improved compared to the single forcing, the modified Transformer was able to utilize the short-term and long-term dependencies of multiple forcing datasets. For one particular ensemble member (random seed), the cumulative density plot shows very similar curves between the modified Transformer and LSTM.

The high agreement between the two model architectures, both of which are state-of-the-art, suggest that we are likely at or very close to the predictive limit of the CAMELS dataset for this test (temporal test). We suspect, unless we bring in new information, it is highly unlikely for other models to produce noticeable advantages beyond these two models on this dataset, for the tests presented here. Errors with forcing, attribute and discharge data would prevent high performance. It should be mentioned that for another test, e.g., prediction in ungauged regions or spatial extrapolation test, physics-informed hybrid models (called differentiable models) can actually outperform LSTM (Feng, Beck, et al., 2022; Feng, Liu, et al., 2022; Tsai et al., 2021). Moreover, several issues surrounding the CAMELS dataset include using basin-average attributes that cannot resolve subbasin-level spatial heterogeneity, using daily precipitation that does not represent hourly rainfall intensity, a fraction of basins with major reservoirs, and the inclusion some overly large basins.

Nevertheless, exactly because the Transformer does not have time integration, it can be trained in a highly parallel fashion and is suited to learning from large amounts of data. As the amount of data and the amount of neurons increase, it is possible to observe emergent behaviors of intelligence (Bubeck et al., 2023). This is a property that is worth further exploring in future studies in hydrology and geosciences. We leave to future work how to incorporate more data with the modified Transformer and test the model for spatial extrapolation (under data-scarce scenario) and temporal extrapolation (for multidecadal projection of trends).

## *Conclusion*

As an initial step, we compared a vanilla Transformer encoder and a modified Transformer to the current state-of-the-art model, LSTM, on the CAMELS benchmark. However, the vanilla Transformer seems to miss some critical functionality so that it is not optimal for simulating discharge. The modified Transformer with no recurrent connection obtains essentially the same metric (albeit only with a slight advantage) as LSTM and works. This means we can continue to search for better architecture to further improve its performance and suitableness for natural physical systems. Our current setup may not be optimal yet. Nevertheless, the differences are overall small between the models, suggesting that we are already close to the optimum for this dataset and this test and more expansion of dataset will be required. We argue the modified Transformer is a viable alternative to LSTM and may find advantages for future, larger datasets.